# Local image registration a comparison for bilateral registration mammography

José M. Celaya-Padilla *[a], Juan Rodriguez-Rojas [a], Victor Trevino [a],
José G. Gerardo Tamez-Pena [a]

[a] Instituto Tecnológico y de Estudios Superiores de Monterrey, Eugenio Garza Sada 2501 Sur Col. Tecnológico C.P. 64849, Monterrey, Nuevo León, México

## ABSTRACT

Early tumor detection is key in reducing the number of breast cancer death and screening mammography is one of the most widely available and reliable method for early detection. However, it is difficult for the radiologist to process with the same attention each case, due the large amount of images to be read.  Computer aided detection (CADe) systems improve tumor detection rate; but the current efficiency of these systems is not yet adequate and the correct interpretation of CADe outputs requires expert human intervention. Computer aided diagnosis systems (CADx) are being designed to improve cancer diagnosis accuracy, but they have not been efficiently applied in breast cancer. CADx efficiency can be enhanced by considering the natural mirror symmetry between the right and left breast.  The objective of this work is to evaluate co-registration algorithms for the accurate alignment of the left to right breast for CADx enhancement.  A set of mammograms were artificially altered to create a ground truth set to evaluate the registration efficiency of DEMONs , and SPLINE deformable registration algorithms. The registration accuracy was evaluated using mean square errors, mutual information and correlation. The results on the 132 images proved that the SPLINE deformable registration over-perform the DEMONS on mammography images.

## 1. INTRODUCTION

Breast cancer is one of the leading causes of women death all over the world [1]. The most effective modality for detecting the early-stage breast cancer is screening mammography, which is an inexpensive tool to classifying abnormalities such as calcifications and masses, as well as subtle signs, like architectural distortion and bilateral asymmetry. In recent years there has been a great effort in research to develop computer-aided detection or diagnosis (CAD) systems that use computer technologies to assist radiologists in their decisions were a high number of studies must be examined.

The detection of these abnormalities in mammograms is commonly performed by comparing the images of the same patient, in either the same breast at different visits, or using the left and right breast (bilateral comparison) [2], however, because the breast is a highly dynamic organ the comparison is not straightforward. There are several challenges that the radiologist need to deal with: the breast can change in size and shape, differences in breast compression, scanner noise etc. Therefore, in order to efficiently compare the breast, an alignment (registration) of the breast must be performed prior to the study [3].

Image registration is used to find an optimal geometric transformation between corresponding images and for this work we can classify it as "rigid" transformation (where the image only need to rotate or been translated to match the image target), or "non-rigid" registration [4,5]  where due the nature of the breast, some local and global transformation such as stretching, compression, rotation etc. are applied to the breast to match the target. The nature of breast imaging requires non-rigid registration, or an hybrid combination of both. Most of the approaches in breast registration use images features such as breast boundary information, and internal regions to obtain a more robust registration. Other approach is the intensity based [6], where they match the intensity patterns in the images. Commonly a measure of similarity that weights how well the registration was done is defined. The transformation that is being applied to the image is adjusted until the similarity measurement is optimized.

The objective of this work is to review of two intensity based registration algorithms tailored to do bilateral breast registration of mammographic images, that can help the CADx and the radiologist to improve diagnosis, this paper is written as follows: in section II we present the methodology of the work, we briefly explain the algorithms used, the dataset and the metrics used for the evaluation of the algorithm. In section III, we present the results and the details obtained. Finally in section IV the discussion and conclusions are presented.

## 2. METHODOLOGY

In order to analyze the performance of the bilateral registration, we follow the process in figure (1). This can be grouped in three main stages, first we develop a synthetic images data set from a previously classified normal bilateral cranio-caudal (CC) and bilateral medium-lateral oblique (MLO), then the registration framework is applied, the last stage is the metrics quantification to evaluate the performance of the methods, the detailed process of each stage is explained above.

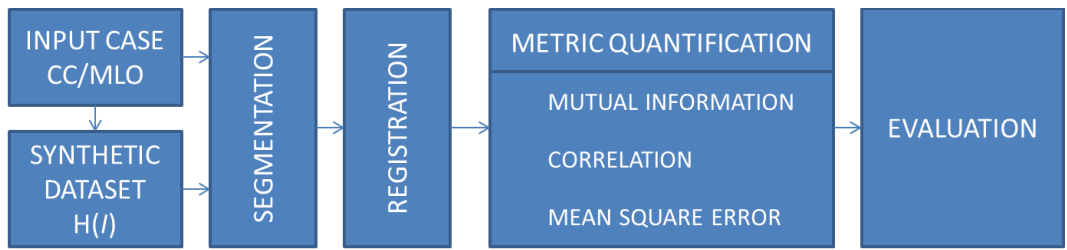

Figure 1. Registration framework process

### 2.1 Synthetic dataset

There are a few data sets of images typically used to test CADx systems, however many data set suffer from the heterogeneity present in the images, while this is a good factor for a more robust classification system, in this approach, we try to limit the effect of noise present in the dataset itself. In order to avoid it, we created a synthetic image dataset in which we will have the control of the modifications and of course the ground truth, this approach is not new, many registration methods are compared with synthetic images such as geometric figures, or a set of specific cases for the field of interest [3], nevertheless our approach try to test the capabilities in a real world scenario, and bilateral registration have special problems, noise, high contrast, high dense areas etc., therefore we created a synthetic dataset based in selected cases of the Digital Database for Screening Mammography (DDSM)[8], the original dataset is a set of mammograms in a digital format that has been used by researchers to evaluate and compare the performance of computer-aided detection (CADx), from this dataset seed, we selected a set of images that were a typical case from three diagnosis type, normal, calcification and masses, those potential cases, were used as a seed to create a new dataset.

The new dataset was created by taking in consideration the common alteration that a mammography may suffer in screening procedure, or by the nature of the breast, those alteration are typically calcifications, and masses found in one breast and not in the other or in a bigger proportion, and the asymmetry of the breast itself.

The process to create the data set was as follow, from a normal classified images, a series of modifications were performed, first lesions were added, for both calcifications and masses, next the following transformations were applied: rotation, compression, deformation, expansion, those transformations were implemented using only global transformations such as affine and Rigid, in order to preserve the changes along the entire image, a total of 132 images were developed, 33 for RCC, 33 for LCC, 33 for LMLO, and 33 for RMLO, plus the RCC, LCC, LMLO, RMLO ground truth, the size was cropped to a resolution of 4376 x 2728 for all the images, 16 bits and keep the original 0.05/0.05 spacing, finally were exported to a friendly DICCOM format.

The next step was the segmentation of the background and unwanted information, the first step is to segment the tissue from the right and left images to eliminate the background and artifacts as name tags and annotations, and to define region of interest (ROI) where the registration and the metrics will be applied, this operation was achieved by using a connected components segmentation algorithm with a 3 pixel radius and a threshold estimated of 12 standard deviations.

## 2.2 Registration framework

We perform the registration using two common used algorithms DEMON's [7] and multi resolution B-Spline[8] in two ways, first we register the transformed image *H(I)* against his own view, and then in order to consider bilateral differences we register the transformed view against his opposite view (e.g. right CC to left CC) , the first registration, give us the ground truth in how the algorithm performs with only transformations, and the second registration give us the relative performance in common scenarios where the target is more complex to achieve.

## 2.3 DEMONS

The Demon's algorithm[7] is based on Maxwell paradox, making reference to those *Demons* in which every pixel in the reference image is applying a *force* in the moving image by pushing the contours of the image in the normal direction of the fixed image, the demons algorithm relies on the assumption that pixels representing the same homologous point on an object have the same intensity on both the fixed and moving images to be registered in (1),

This algorithm handles the registration as a diffusion process, where the diffusivity is related to the local characteristics of the region to be register. This method does not minimize a global objective function; it works locally following the optical flow principle, using a deformation field, the algorithm was made using the Thirion et al [7][9] implementation in the Insight Toolkit (itk) libraries [10].

$$D^N(X) = D^{N-1}(X) - \frac{\left(m\left(X+D^{N-1}(X)\right)-f(X)\right)\nabla f(x)}{\|\nabla f\|^2 + \left(m\left(X+D^{N-1}(X)\right)-f(X)\right)^2} \quad (1)$$

In the above equation, D(X) is the displacement or optical flow between the images, f (X) is the fixed image, m(X) is the moving image to be registered [7][9].

We also use a variation of the DEMON's algorithm, which takes in account both gradients of the fixed and moving image in order to compute the deformation forces. This mechanism for computing the forces introduces symmetry with respect to the choice of the fixed and moving images *k* in (2)

$$D(X) = - \frac{2\bigl(m(X)-f(X)\bigr)\bigl(\nabla f(X)+\nabla g(X)\bigr)}{\|\nabla f+\nabla g\|^2+\bigl(m(X)-f(X)\bigr)^2/K} \quad (2)$$

The framework for DEMON's registrations was as follow: first the moving image (image to be register) is swapped if needed, then a down sampled from both the moving image and the target image is generated at a resolution of 219x136, then a histogram matching is applied to ensure that both images have the same scale of intensities, afterward registration is performed on the down sampled images, and the deformation field *D(I)* (equation 1) is calculated, after this the deformation field *D(I)* is up sampled to the original dimensions to match the size, finally the original moving image is warped using the deformation field to generate the registered image, the set of parameters were obtained by choosing the best set of parameters with the best result for all the different images.

## 2.4 B-splines

The B-Spline[8] algorithm is based on deforming an image by modifying a mesh of control points following a maximization of a similarity measure[11]. The degree of deformation of the mesh can be adjusted with the resolution of the mesh, for this algorithm we follow the multi resolution approach, similar to the demons approach, first the images are registered in a lower resolution, propagating parameter estimation into a higher resolution and performing registration again, this often avoids local minima in the parameter search space and reduces computational time [8].

One key difference from the previous algorithm is that this approach tries to minimize a global objective, or measurement of the fitness in the overlap of the images, for this algorithm the framework was as follow: first the moving image (image to be register) is swapped if needed, then a down sampled from both the moving image and the target image is generated at a resolution of 219x136, then the pyramids for the multi resolution are generated, afterwards registration is performed, then the original moving image is deformed using the final parameters of the registration and a upscale transformation, in this implementation, a mutual information metric was used as a metric for optimization, implementation was made based in the Insight Toolkit (itk) libraries [10], the set of parameters were obtained by choosing the best set of parameters with the best result for all the different images.

## 3. METRIC QUANTIFICATION

Intensity based registrations match intensity patterns over the image, however this type of registration does not use anatomical knowledge, therefore an assumption is made, that the images will be most similar between them when the registration is maximized, the most common measurements of similarity include: sum of square differences, correlation coefficient, and information-theoretic such as mutual information.

The sum of squared differences (SSD) in (3) assumes that the images are identical at registration except for (Gaussian) noise, this metric is sensitive to small number of voxels that have very large intense differences, this metric is recommended for mono-modal registration only, *f(x)* is the intensity at a position x in an image, *m(t(x))* is the intensity at the corresponding point given by the transformation *t(x)*. *N* is the number of pixels in the region of overlap. [12]

$$SSD = \frac{1}{N}\sum_X \left(f(X) - m(t(X))\right)^2 \tag{3}$$

The SSD measure makes the implicit assumption that after registration, the images differ only by Gaussian noise. A slightly less strict assumption would be that, at registration, there is a linear relationship between the intensity values in the images. In this case, the optimum similarity measures is the correlation coefficient CC in (4) where *f(x)* is the intensity at a position *x* in the fixed image, *m(t(x))* is the intensity at the corresponding point given by the transformation *t(x)* in the moving image. *N* is the number of pixels in the region of overlap.

$$CC = \frac{\sum_X (f(X)-\bar{f})*(m(t(X))-\bar{m})}{\sqrt{\sum_X (f(X)-\bar{f})^2 * \sum_X (m(t(X))-\bar{m})^2}} \qquad (4)$$

Another well know similarity measurement is mutual information in (6), that provides a measure of probabilistic mutual dependence between two intensity distributions, The most commonly used measure of information in image processing is the Shannon–Wiener entropy measure *H* in (5) , originally developed as part of communication theory in the 1940s (Shannon 1948, 1949) [13], a common approach larger MI means more similar images are.
.

$$H(X,Y) = -\sum_{x,y=0}^{N} P_{x,y} \log_2(P_{x,y}) \qquad (5)$$

$$H(X) = -\sum_{x=0}^{N} P_x \log_2(P_x)$$

In (5) *H(X,Y)* represents the joint entropy, and *H(X)* the individual entropy of random image *X, Y, N* stands for the number of intensity levels and *Px (Pxy )* is the probability of value *X (x,y)* in the (joint) probability distribution of variable *X (X and Y )*.

$$MI(m,f) = H(m) - H(f|m) \qquad (6)$$

Image processing where performed using a Core 2 duo computer using windows 7, C++ and using the software CiPAS v1.0 [14] (IMITEK, Monterrey, Mexico) and libraries from the Insight Toolkit (ITK)[10].

## 4. RESULTS

In figure 2 we can observe a registration result, for both the B-Spline and DEMON's registration, as we can see the algorithms achieve the target and perform very well, in figure 3 (a), is clear the impact of the registration, the difference of the ground-truth against both registered cases achieve more MI, for the bilateral registration the result is similar, however the DEMON's variant achieve a relative better result.

For every scenario B-Spline perform very well, and overcome DEMON's in some scenarios, however we can notice that inside the breast, booth DEMON's algorithm introduced artifacts in some cases, in which the B-Splines does not and have stable behavior in all the tests where the DEMONS trend to vary along, further more B-Splines can register the image in no more than 100 iterations, where DEMON's needs between 300 - 1000 iterations to complete the registration, we assume that the difference in the speed came from the implementation of the B-spline, the pyramids multi resolution implementation reduce the amount of compute time, for the correlation and mean squares metrics we notice similar performance to the mutual information presented in figure3.

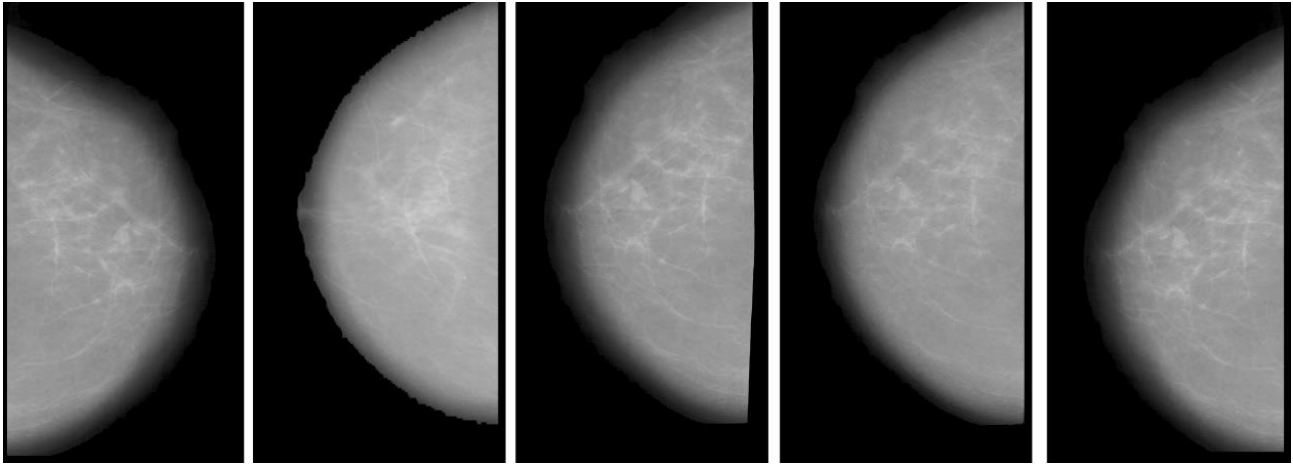

Figure 2. a) LCC Target image and segmented RCC image to be register, c) Demon's result c) Demon's gradient variant result d) B-Splines result.

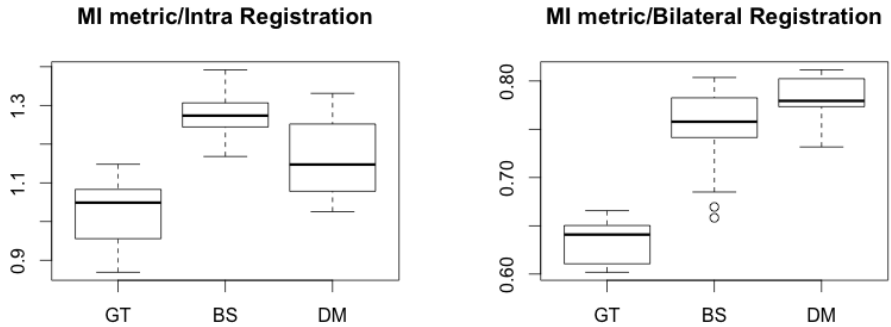

Figure 3. Mutual information (MI) boxplot, a)Mutual information metric, for the intra modal registration, GT = ground truth the metric obtained without registration, BS =B-Splines registration result, DM = DEMON's gradient variant registration result, b) Mutual information metric, for bilateral registration, GT = ground truth the metric obtained without registration, BS = B-Splines registration result, DM = DEMON's registration result.

In table 1, we present a snap shot of the metrics gathered in the different post bilateral registration experiments, we group the information in normal, masses, and calcification cases, the information is presented in the difference observed against the ground-truth of each metric, here we can see a big difference in the SSD error with an average improve of 44,744.04, 8401.69, 143,804.39 for (B-splines, DEMON's. DEMON's gradient variant) this metric shows a huge improvement after registration. The demons variant shows a better performance, the internal deformations can lead to a unrealistic transformations, and the internal structure of the breast can be compromise due this deformations, due the big size of the dataset is normal that the numbers shown above were large, in contrast in the correlation metric there is not a huge difference, this due the fact that the ground-truth metric showed a big correlation since the beginning and there wasn`t much work to do there, and last, for the mutual information, is clear the improvement over the ground-truth, in all the metrics an average improve of 0.6, 0.52, 0.74 for (B-splines, DEMON's. DEMON's gradient variant).

Table 1. Bilateral registration metric differences against the ground truth-value, for mutual information and correlation more means better alignment, for sum of square differences less mean better alignment.

| Classification | Mutual Information | | | Correlation | | | SSD | | |
|---|---|---|---|---|---|---|---|---|---|
| Normal | B-Spline | Demons | Dem G | B-Spline | Demons | Dem G | B-Spline | Demons | Dem G |
| Compression | 0.6497 | 0.5575 | 0.7778 | 0.0349 | 0.0222 | 0.0535 | -50548.85 | -4847.04 | -146694.20 |
| Movement | 0.6467 | 0.5024 | 0.7101 | 0.0330 | 0.0226 | 0.0523 | -39378.64 | -5259.96 | -140275.88 |
| Deformation | 0.2115 | 0.5188 | 0.7409 | 0.0341 | 0.0225 | 0.0530 | -46000.37 | -5188.47 | -144253.64 |
| Masses | | | | | | | | | |
| Compression | 0.6514 | 0.5501 | 0.7714 | 0.0347 | 0.0224 | 0.0538 | -49327.79 | -5610.17 | -147860.82 |
| Movement | 0.6476 | 0.5018 | 0.7078 | 0.0327 | 0.0229 | 0.0522 | -37869.47 | -6695.23 | -139837.57 |
| Deformation | 0.6290 | 0.5191 | 0.7389 | 0.0341 | 0.0223 | 0.0530 | -45554.29 | -4427.99 | -143860.65 |
| Calcification | | | | | | | | | |
| Compression | 0.6497 | 0.5456 | 0.7620 | 0.0344 | 0.0236 | 0.0532 | -47353.42 | -10671.37 | -144798.19 |
| Movement | 0.6526 | 0.5215 | 0.7313 | 0.0334 | 0.0235 | -0.1130 | -41866.85 | -36029.53 | -143416.78 |
| Deformation | 0.6346 | 0.5037 | 0.7228 | 0.0339 | 0.0206 | 0.0529 | -44796.65 | 3114.56 | -143241.78 |

In figure 3, we plot the joint entropy histogram (JEH) of a sample case, this is figure shows the comparison of joint entropy histogram of the images, the basic idea is that when the images are correctly registered, the joint histograms have tight clusters, surrounded by large dark regions and these clusters disperse as the images become less well registered, in this similarity measurement it is clear that the best result is achieved by the B-spline registration, the clusters are very concentrated in a very stretch zone, on the other hand, for the DEMON's algorithm is clear that the clusters are dispersed and lead us to infer that the registration was not good performed, the gradient variation of the DEMON's algorithm, achieved an overall best result the cluster are more compacted but not as good as the B-spline result.

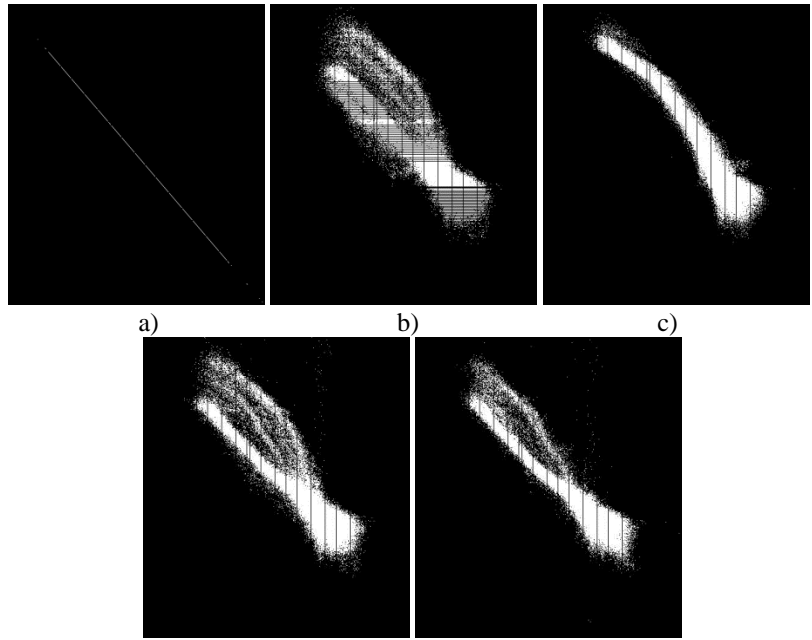

a)                b)                c)

d) e)

Figure 3. Joint Entropy Histogram (JEH) of a) identity transformation (same image), b)No registration Left CC view to Right CC view, c)B-spline Left CC view to Right CC view registration  d) Demons Left CC view to Right CC view registration, e) Demons gradient Left CC view to Right CC view registration.

## 5. CONCLUSIONS

We presented a methodology to compare two local approach registration methods to help CADx systems to perform morphing operations, our work differ from other comparisons of register algorithms, in two key aspects, 1) we developed the Synthetic data set where previous works only test in public heterogeneous datasets, in which the transformations are unknown 2) Our approach used a tissue mask for the evaluation of registration accuracy.

In this work we developed a methodology to perform automated registration comparison for bilateral mammography, because most current algorithms for medical image registration calculate a rigid body or affine transformation, their applicability is restricted to parts of the body where tissue deformation is small compared with the desired registration.

The B-spline algorithm achieve an overall better results, in some conditions, DEMON's shows a better performance, however the approach of this algorithm can lead to a unrealistic transformations, which in practice achieve a better final similarity measurement, but in real life, the deformations could lead to a un unrealistic image deformations for the radiologist usage.

The joint entropy histogram of the B-spline algorithm shows a better clustering, while booth DEMON's algorithms the histogram is more scattered across the plane, which suggest a better alignment for the B-Spline, while keeping the information contained inside the breast.

The performance of the algorithms DEMON's and B-spline was similar. The correlation doesn't show a big improvement in the registration. However the fact that the correlation uses the whole image obtain the metric, this show a high correlated image, but nevertheless there was an improvement, the SSD metric was the metric that had the biggest improvement over the ground-truth in all scenarios and in all types of register methods, the MI metric shown the improvement against the grown-truth with an average improvement of 0.6, 0.52, 0.74 for (B-splines, DEMON's. DEMON's gradient variant); but, once again, when we take into consideration the unrealistic transformations that DEMON's algorithm can yield in some scenarios, the B-Spline algorithm should be the right choice.

We demonstrated the performance of the algorithms where the B-Spline algorithm is accurate in almost all scenarios in such a way for the residual image to be used as derivative image to extract more features for a CADx system. In all the registrations we obtain reductions in the metric measurements between images prior and after registrations, Overall we can conclude that B-Splines obtained the best Quantitative and visual results. The work can be improved, by the use of more similarity metrics, such as watershed error, geometric distances etc. and of course a by selecting a larger set of registration algorithms.

## REFERENCES

[1] R. Nishikawa, "Current status and future directions of computeraided diagnosis in mammography," Computerized Medical Imaging and Graphics, vol. 31, pp. 224-235, 2007.
[2] F. Yin, M. Giger, C. Vyborny, K. Doi, and R. Schmidt, "Comparison of  bilateral-subtraction and single-image processing techniques in the computerised
[3] B. Zitova, "Image registration methods: A survey," Image Vision Comput.,vol. 21, no. 11, pp. 977–1000, Oct. 2003.

[4] D. Mattes, D. R. Haynor, H. Vesselle, T. Lewellen and W. Eubank "Nonrigid multimodality image registration" Medical Imaging 2001: Image Processing, 2001, pp. 1609-1620.
[5] D. Mattes, D. R. Haynor, H. Vesselle, T. Lewellen and W. Eubank  "PET-CT Image Registration in the Chest Using Free-form Deformations" IEEE Transactions in Medical Imaging. Vol.22, No.1, January 2003. pp.120-128.
[6] D.L.G. Hill, D.J. Hawkes "Across-modality registration using intensity-based cost functions", I. Bankman (Ed.), Handbook of Medical Imaging: Processing and Analysis, Academic, New York (2000), pp. 537–553
[7] Michael Heath, Kevin Bowyer, Daniel Kopans, Richard Moore and W. Philip Kegelmeyer, "The Digital Database for Screening Mammography", in Proceedings of the Fifth International Workshop on Digital Mammography, M.J. Yaffe, ed., 212-218, Medical Physics Publishing, 2001. ISBN 1-930524-00-5.
[8] J. P. Thirion,  "Fast non-rigid matching of 3D medical images" (1995) Medical Robotics and Computer Aided Surgery, pp. 47-54.
[9] D. Rueckert, L. I. Sonoda, C. Hayes, D. L. Hill, M. O. Leach, and D. J. Hawkes, "Nonrigid registration using free-form deformations: Appli- cation to breast MR images," IEEE Trans. Med. Imaging, vol. 18, no. 8, pp. 712–721, Aug. 1999.
[10] J.P. Thirion. "Image matching as a diffusion process: an analogy with maxwell's demons." MedicalImageAnalysis,2(3):243–260,1998. 8.14
[11] "The Insight Segmentation and Registration Toolkit"   http://www.imitek.com.mx/
[12] Díez, Y.; Oliver, A.; Llado, X.; Freixenet, J.; Marti, J.; Vilanova, J.C.; Marti, R., "Revisiting Intensity-Based Image Registration Applied to Mammography," Information Technology in Biomedicine, IEEE Transactions on , vol.15, no.5, pp.716,725, Sept. 2011
[13] W R Crum, T Hartkens, and D L G Hill "Non-rigid image registration: theory and practice" Br J Radiol December 2004 77:S140-S153; doi:10.1259/bjr/25329214
[14] Shannon, C. E., "A Mathematical Theory of Communication," Bell System Technical Journal
Inter Medical Imaging Technologies CiPAS v1.0  http://www.imitek.com.mx/